\documentclass[11pt,a4paper]{article}
\usepackage[hyperref]{acl2020}
\usepackage{times}
\usepackage{latexsym}

\usepackage{microtype}

\aclfinalcopy 

\usepackage{amsmath}
\usepackage{multirow}
\usepackage{url}
\usepackage{arydshln}
\usepackage{amssymb}
\usepackage{float}
\usepackage{hhline}
\usepackage{graphicx}
\usepackage{float}
\graphicspath{ {./images/} }

\usepackage{url}

\newcommand{\s}{\mathbf{s}}

\DeclareMathOperator{\siml}{sim}
\DeclareMathOperator{\set}{set}
\DeclareMathOperator{\bag}{bag}
\DeclareMathOperator{\preci}{precision}
\DeclareMathOperator{\overl}{overlap}
\DeclareMathOperator{\bleu}{bleu_4}
\DeclareMathOperator{\smoothbleu}{smoothed\_bleu_4}
\DeclareMathOperator{\lstm}{lstm\_states}
\DeclareMathOperator{\transformer}{transformer\_states}
\DeclareMathOperator{\score}{score}

\title{Leveraging Sentence Similarity in Natural Language Generation:\\
{I}mproving Beam Search using Range Voting}

\author{Sebastian Borgeaud\\
  DeepMind \& University of Cambridge \\
  \texttt{sborgeaud@google.com}\\\And
  Guy Emerson \\
  University of Cambridge \\
  \texttt{gete2@cam.ac.uk} \\}

\date{}

\begin{document}
\maketitle
\begin{abstract}
We propose a method for natural language generation,
choosing the most representative output 
rather than the most likely output.
By viewing the language generation process 
from the voting theory perspective,
we define representativeness using range voting 
and a similarity measure.
The proposed method can be applied
when generating from any probabilistic language model,
including n-gram models and neural network models.
We evaluate different similarity measures 
on an image captioning task and a machine translation task,
and show that our method generates longer and more diverse sentences,
providing a solution to the common problem
of short outputs being preferred over longer and more informative ones.
The generated sentences obtain higher BLEU scores, 
particularly when the beam size is large.
We also perform a human evaluation on both tasks and find that the outputs
generated using our method are rated higher.
\end{abstract}

\section{Introduction}
\label{sec:intro}

A language model specifies a probability distribution over sequences of words:
given a sequence~${s=x_1 x_2 \cdots x_n}$ of length $n$,
the model assigns a probability~$P(s)$ to the entire sequence.
The probability distribution may be conditioned:
for example in machine translation the distribution is conditioned on the source language sentence. 

In many applications, it is desirable to output a single sequence, rather than a distribution.
A common approach is to choose the most likely sequence.
However, this is problematic when the most likely sequence
is not \textit{representative} of the whole distribution.

For example, in dialogue generation tasks, the most likely output can be ``I don't know'',
even when most of the probability mass is assigned to long informative sequences.
\citet{cao2017latent} call this the ``boring output problem''.

For a real-valued distribution, we can choose a representative output by taking the mean.
However, for a discrete distribution (such as over sequences), the mean is not defined.
In this paper, we choose a representative output using tools from voting theory,
allowing us to avoid the boring output problem.
The general idea is that,
if the distribution assigns most of the probability mass to a group of similar sequences,
we would like to generate one of them
-- even if they have low probability as individual sequences,
they have high probability as a group.
We can formulate this process as a range voting election,
where the sentences vote for each other,
with the strength of a vote being proportional
to the similarity between the voter sequence and the candidate sequence.

Our approach can be used to mitigate
problems commonly associated with language models.
For example, a long-recognised problem is 
that shorter sequences are assigned higher probabilities 
and thus choosing the most likely sequence favours short sequences \citep{brown1995method}.
Indeed, \citet{stahlberg2019nmt} show that the most likely 
output in machine translation is often the empty string.
By designing the similarity function to be asymmetric 
such that more informative candidate sequences receive stronger votes,
we can generate longer and more diverse outputs
(see Fig.~\ref{image/example_image} for an example).

We focus on simple similarity metrics based on n-grams 
and generate the candidates and voters using beam search.
We evaluate on two tasks: image captioning and machine translation.
For both tasks,
we find that our approach achieves higher BLEU scores,
and performs better in a human evaluation.
Our approach also generates longer and more diverse outputs,
with the generated length and diversity more closely matching
the length and diversity of the reference captions
and reference translations.

\begin{figure}
\vspace*{1mm}
\begin{minipage}{0.25\columnwidth}
	\includegraphics[width=1.0\columnwidth]{6843}
\end{minipage} \hfill
\begin{minipage}{0.72\columnwidth}
	\scriptsize{
	0.00230: a couple of people that are sitting on a bench\\
	0.00132: a man sitting on a bench next to a dog\\
	0.00079: a black and white photo of a man sitting on a\\\hphantom{0.00079: {}}bench\\
	0.00075: a couple of people sitting on a bench\\
	0.00066: a man sitting on a bench with a dog\\
	0.00064: a man and a woman sitting on a bench\\
	0.00048: a man and a woman sitting on a park bench\\
	0.00046: a black and white photo of a man and a horse\\
	0.00033: a black and white photo of a man and a dog\\
	0.00025: a black and white photo of a man on a horse
	}
\end{minipage}
\caption{%
	Image from the MSCOCO validation dataset
	and beam search captions with their probabilities
	(beam size $k{=}10$).
	See \S\ref{sec:beam} for beam search, \S\ref{sec:image_captioning} for the sequence model.
	Range voting with $\overl_2$ similarity (see \S\ref{sec:vote})
	on this set of sequences
	selects\linebreak``a black and white photo of a man sitting on a bench'',
	which shares many bigrams with other sequences.
}
\label{image/example_image}
\end{figure}

\section{Related work}
\label{sec:relwork}

Much work has gone into analysing sources of errors in language generation,
often focused on machine translation.
\citet{koehn2017six} raise 6 challenges for machine translation, 
including degrading performance for longer sentences,
and degrading performance for larger beam sizes.
\citet{stahlberg2019nmt} distinguish
\textit{model errors} (high probabilities of bad sequences)
and \textit{search errors} (failing to find sequences preferred by the model).
They show that the global optimal translations 
(according to likelihood) 
are considerably worse than translations found by beam search.
This points to both serious model errors and serious search errors,
which cancel out to some degree.
This suggests there is much work to be done
in improving both our models and our search objectives
-- the latter is the aim of this paper.

\citet{ott2018} find that beam search typically covers
only a small proportion of the model's probability mass,\footnote{%
	Since our paper was submitted,
	this finding was replicated by \citet{eikema2020mode},
	who further argue that the maximum-likelihood decoding objective
	is hard to justify when the maximum likelihood is so low.
}
and they show that the degradation for large beams is
at least partly due to the training data
containing target sentences that are exact copies of source sentences.
They also suggest that beam search is an effective search strategy,
for the maximum-likelihood search objective,
finding hypotheses with higher model probabilities than the reference translations.

\citet{pmlr-v97-cohen19a} also find a performance degradation with larger beam sizes 
across different tasks (translation, image captioning and summarisation)
and propose to add a search discrepancy heuristic to beam search.
For image captioning, \citet{VinyalsShowAndTellLessons2017} show that 
larger beams not only decrease performance but also reduce the diversity of the captions.
They claim this is an overfitting effect and propose the use of small beam sizes as further regularization.

In unconditional, open-ended language generation, 
\citet{holtzman2020curious} find that using likelihood as the decoding objective
leads to bland and repetitive text with unnaturally high probability and too little variance. 
They claim this is not due to a search error,
but due to the maximum-likelihood decoding objective.
They propose sampling,
truncated the distribution to the top~$p$ percent of tokens.

\subsection{Generation length and diversity}
\label{sec:relwork:length}

To increase the length and diversity of a model's outputs,
some authors have proposed changes to the model architecture.
In dialogue generation, \citet{cao2017latent} use a latent variable model
to capture the possible ``topics'' of a response.

Others have proposed changing the objective function.
In dialogue generation, \citet{li2015diversity} optimise mutual information instead of probability.
In machine translation, \citet{tu2017neural} modify an encoder-decoder model
by adding a ``reconstructor'' to predict the input based on the output.

However, modifying the model or the objective function depends on the particular task,
and applying these techniques to an existing system requires retraining the model.
In this paper, we focus on general methods which can be applied
to any probabilistic model in any generation task.
Length normalisation \citep{wu2016google,freitag2017beam} explicitly penalises shorter sequences during the beam expansion phase by dividing the log-probability of a sequence by its length. 
Diverse decoding \citep{li2016simple, li2016mutual} penalises repeated expansions of the same beam node.
Diverse beam search \citep{diversebeam-vijay} penalises generation of similar beams using their Hamming diversity.
These last two methods aim to increase the diversity within a beam, but not necessarily across the dataset.

\citet{kool2019stochastic} propose a stochastic beam search 
based on the Gumbel-Top-$k$ trick to sample without replacement.
The proposed approach can trade-off BLEU score against translation diversity.

Finally, it is important to make sure that
improvements to a model can be properly evaluated.
After our paper was submitted, \citet{freitag2020bleu} report that the references used in machine translation often exhibit poor diversity,
which can unfairly penalise models which exhibit good diversity.
They propose to use paraphrased reference translations instead.
These paraphrases yield higher correlation with human judgement when evaluated using BLEU,
and could be used in future work to improve the evaluation of translation systems which aim to generate appropriately diverse outputs.

\subsection{Minimum Bayes Risk Decoding}
\label{sec:relwork:mbr}

\citet{kumar2004mbr} introduce the Minimum Bayes Risk (MBR) decoder for machine translation.
Like our proposed approach, this aims to use the whole distribution,
rather than picking the most likely sequence.
They frame the problem in terms of Bayes Risk:
given the true distribution over outputs,
and given a loss function between the system output and the target output,
the Bayes Risk is defined as the expected loss.
The best output is the one which minimises the Bayes Risk.

However, the true distribution over outputs is not known,
so \citeauthor{kumar2004mbr} approximate it using the model's distribution.
The MBR decoder first uses beam search,
and then re-ranks it according to the BLEU scores between sequences in the beam.

\citet{tromble2008lattice} apply MBR over translation lattices.
\citet{shimizu2012minimum} use MBR with a smoothed BLEU loss
function and propose to limit the possible translations
to those that are similar to most-likely translation generated 
by beam search. 

\citet{blain2017exploring} propose to re-rank the sentences generated 
by beam search using a similarity metric.
Their approach is similar to ours but doesn’t include the probability 
of the sentences given by the decoder,
and thus would degrade completely 
in the limit of very large beam sizes.
They find that using BLEU as a similarity metric reduces the 
quality of generated translations,
according to both BLEU and a human evaluation.

\section{Method}
\label{sec:method}

\subsection{Beam search}
\label{sec:beam}

When working with a distribution over sequences,
it is not feasible to consider all possible sequences.
Finding the most likely sequence can be computationally expensive
-- in fact, for an RNN it is undecidable
in the general case \citep{chen2017recurrent}.
A common solution is to use \textbf{beam search},
which generates the sequence one token at a time,
maintaining a list of the $k$~most promising sequences at each time step
(for example:\ \citealp{brown1995method,koehn2004pharaoh}).
\textbf{Greedy search} is the special case where $k=1$.

Beam search introduces an extra hyper-parameter, the beam size~$k$.
Increasing~$k$ covers more of the search space, but increases the computational cost.
It is tempting to assume that increasing~$k$ will produce better results,
but empirically, the quality of the most likely sequence
starts to decrease after~$k$ exceeds a certain threshold \citep{koehn2017six, pmlr-v97-cohen19a}.

In the next section, we propose an alternative way to generate from a beam,
which aims to avoid the drop in performance as beam size increases.
Rather than choosing the most \textit{likely} sequence,
we choose the most \textit{representative} sequence.

\subsection{Range voting}
\label{sec:vote}

To formalise representativeness,
we propose to use a voting procedure.
Although voting has been applied to ensembles of classifiers
(for an overview, see:\ \citealp{kuncheva2004combining,kuncheva2014ensemble}),
we are not aware of work using voting to select from a distribution.

We can see each sequence as a candidate in an election,
and the probability of a sequence as the proportion of votes for that candidate.
From this perspective, the problem of probability mass being split across long sequences
is the well-known problem of \textbf{vote splitting}.
Suppose candidate~$i$ wins an election.
Now suppose we run the election again,
but add an additional candidate~$j$, identical to~$i$.
A voting system is robust against vote splitting (and called \textbf{independent of clones})
if the winner must be $i$~or~$j$ \citep{tideman1987clone}.

A well-studied system which is independent of clones is \textbf{range voting}
\citep{heckscher1892,smith2000range,tideman2006,lagerspetz2016}.
Each voter scores each candidate in the range~$[0,1]$,
and the candidate with the highest total score wins.

In our setting, probability mass can be seen as the proportion of votes placing a candidate as first choice
(see Fig.~\ref{image/example_image} for an example).
For range voting, we need to augment the votes with scores for all other candidates.
We propose to do this using a similarity measure.
The final score for a sequence $\mathbf{c} \in \mathcal{C}$ (the set of candidates) is given in~(\ref{eqn:score}),
for a set of voter sequences~$\mathcal{V}$
and a similarity measure~$\siml$.
\begin{equation}
	\score(\mathbf{c}) = \sum_{\mathbf{v} \in \mathcal{V}} P(\mathbf{v}) \cdot \siml(\mathbf{v}, \mathbf{c})
\label{eqn:score}
\end{equation}

A sequence can act as both voter and candidate.
Each voter sequence is weighted by its probability,
and casts a vote for each candidate sequence,
where the strength of the vote is the similarity between the voter and the candidate.
The simplest way to apply this method
is to use beam search
to define both the set of candidates and the set of voters.

This can be seen as a generalisation of taking an average.
In a Euclidean space,
the mean is equivalent to voting with quadratic similarity~${1-k(x-y)^2}$,
and the median is equivalent to voting with linear similarity~${1-k|x-y|}$,
for some constant $k$.

Although the vote splitting problem may appear abstract, 
it can happen in practice, even without considering similarity.
When using subword vocabularies \citep{sennrich2016neural},
there are multiple ways of encoding any given sentence.
The model's probability mass is split across
sentences with identical surface form but different encodings.

Defining semantic similarity between sentences
is recognised as a hard problem \citep{achananuparp2008sim,cer2017semeval,pawar2019sim}.
In this work, we focus on simple, domain-agnostic similarity measures
which do not require additional training.

First, we consider similarity based on n-grams.
For a sequence~$\s$,
we write $\set_n(\s)$ for its set of n-grams,
and $\bag_n(\s)$ for its bag (or multiset) of n-grams.
We define two measures \mbox{in~(\ref{eqn:preci}--\ref{eqn:overl})}.  
Both are asymmetric, to encourage informative sequences:
if $\mathbf{c}$~contains $\mathbf{v}$ plus more information,
$\siml(\mathbf{v},\mathbf{c})$~should be high,
but if $\mathbf{c}$ contains less information, then $\siml(\mathbf{v},\mathbf{c})$~should be lower.
This allows an informative candidate sequence to gather more votes.
\begin{align}
	\preci_n(\mathbf{v},\mathbf{c}) &= \frac{| \bag_n(\mathbf{v}) \cap \bag_n(\mathbf{c}) |}{|\bag_n(\mathbf{v})|} \label{eqn:preci}\\
	\overl_n(\mathbf{v},\mathbf{c}) &= \frac{| \set_n(\mathbf{v}) \cap \set_n(\mathbf{c}) |}{|\set_n(\mathbf{v})|} \label{eqn:overl}
\end{align}

Second, inspired by \citet{mueller2016}, we consider a similarity measure based on 
the hidden states of the decoders (LSTM and Transformer) during generation (see~\S\ref{sec:image_captioning}).
For each sequence, we find the average of the hidden states,
and then compute the cosine similarity.
We refer to this measure as $\lstm$ and $\transformer$.


\subsection{Comparison with MBR Decoding}

The formulation used for range voting is reminiscent of MBR decoding
(see~\S\ref{sec:relwork:mbr}).
In fact, if the similarity measure in~(\ref{eqn:score})
is $\siml(\mathbf{v}, \mathbf{c}) = \textrm{BLEU}(\mathbf{c}, \mathbf{v})$,
range voting recovers MBR decoding.
From a theoretical point of view,
range voting provides an independent motivation for MBR decoding,
and furthermore, one which does not require the assumption that
we can approximate the true distribution by the model's distribution.
We know that the model's distribution does not match the true distribution
(or else we would have already solved the task),
and so this is a strong assumption to make.

From a practical point of view,
range voting suggests that any similarity measure could be used,
and not necessarily the evaluation metric.
Using BLEU has several disadvantages.
Firstly, BLEU can be harsh:
when there are no 3- or 4-gram matches, the score is 0.
Secondly, BLEU is a corpus-level metric which does not decompose over sentences.
Finally, BLEU is precision-based,
penalising translations containing information that is not in the reference.
In MBR, this means that candidate sequences are penalised
for containing more information than voter sequences.
Our proposed similarity measures are asymmetric in the opposite direction,
to encourage generation of long and informative sequences.

Indeed, in our experiments,
we have found that simple similarity measures
produce longer and more diverse sentences than BLEU,
and for translation better results,
even though BLEU is used as the evaluation metric.

Furthermore, the voting theory perspective
can yield analytical insights even when range voting is not used.
For example, the performance degradation found by \citet{pmlr-v97-cohen19a}
can be interpreted in terms of vote splitting.
They argue for the need to filter out
sequences which begin with a low-probability token
that is followed by very-high-probability tokens,
in favour of sequences where all tokens have fairly high probability.
The sequences they want to filter out have \emph{not} split the vote
(later tokens have probability close to~1,
so there are no similar sequences that have high probability),
but the sequences they want to keep \emph{have} split the vote
(there are similar sequences with similar probability).
Their method aims to remove these problematic sequences that don't split the vote,
while our method aims to be robust against vote splitting.

\section{Experiments}
\label{sec:exp}

\begin{table*}
	\centering	
	\begin{tabular}{l  c  c  c  c | c  c  c  c}
		& \multicolumn{4}{c}{BLEU-1} & \multicolumn{4}{c}{BLEU-4}\\ 
		Beam size $k$ & 1 & 2 & 10 & 100 & 1 & 2 & 10 & 100\\
		\hhline{=========}
		Beam search & 
		66.66 & 67.97 & 67.22 & 66.18 & 
		25.39 & 26.83 & 27.16 & 26.31 \\
		\hdashline
		Length normalisation & 
		66.66 & 68.47 & 64.72 & 63.10 & 
		25.39 & 26.72 & 25.76 & 24.72 \\
		Diverse decoding & 
		66.66 & 67.90 & 67.24 & 66.43 & 
		25.39 & 26.68 & 26.93 & 26.37 \\
		\hdashline
		$\overl_1$ & 
		66.66 & \textbf{68.55} & 66.26 & 66.36 & 
		25.39 & 26.47 & 25.61 & 24.60 \\
		$\preci_1$ & 
		66.66 & 68.54 & 66.31 & 66.46 & 
		25.39 & 26.47 & 25.62 & 24.58 \\
		
		$\overl_2$ & 
		66.66 & 68.20 & 67.36 & 67.19 & 
		25.39 & 26.82 & 27.22 & 27.13 \\
		
		$\preci_2$ & 
		66.66 & 68.20 & 67.63 & 67.21 & 
		25.39 & 26.82 & 27.23 & 27.13 \\
		$\lstm$ & 
		66.66 & 67.97 & \textbf{68.42} & 69.10 & 
		25.39 &  \textbf{26.83} & \textbf{27.96} & 28.23 \\
		\hdashline
		$\bleu$ \small{(MBR)} &
		66.66 & 67.93 & 67.66 & 68.56 & 
		25.39 &  26.83 & 27.80 & 28.71 \\
		$\smoothbleu$ {\small{(MBR)}} &
		66.66 & 67.98 & 67.87 & \textbf{69.45} & 
		25.39 &  26.84 & 27.89 & \textbf{29.19} \\
		
	\end{tabular}%
	\caption{%
		BLEU-1 and BLEU-4 scores obtained on the MSCOCO validation images.
	}
	\label{table/bleu}
\end{table*}

We evaluate our method on two tasks: image captioning and machine translation.
For MBR, we use BLEU ($\bleu$) and a smoothed version of BLEU ($\smoothbleu$)
which adds 1 to the $n$-gram counts for~$n{>}1$
to mitigate the harshness of the metric \citep{shimizu2012minimum}.

We consider two baselines: length normalisation and diverse decoding,
described in~\S\ref{sec:relwork:length}. 
For machine translation, we also consider diverse beam search as a further baseline.
Other methods mentioned in \S\ref{sec:relwork}
cannot be straightforwardly applied
as they require modifying the model or the training objective.

\subsection{Image captioning}
\label{sec:image_captioning}

We use the MSCOCO dataset \citep{lin2014microsoft},
which consists of 82,783 training images and 40,504 validation images,
each annotated with 5 captions from human annotators.

We use the ``Show and Tell'' encoder-decoder
architecture of \citet{vinyals2015show}.
The encoder is a pretrained Inception V3 CNN \citep{szegedy2016rethinking}
from which we extract a feature vector from the final pooling layer \citep{ioffe2015batch}.
The decoder is an LSTM \citep{hochreiter1997long} with 512 hidden units,
initialising the hidden state using the encoder.
The vocabulary consists of the 5000 most common words in the training captions,
for which embeddings of size 512 are learned from scratch. 

\subsubsection{BLEU scores}
\label{sec:bleu}
Table~\ref{table/bleu} shows BLEU scores \citep{papineni2002bleu} on the 
MSCOCO validation set computed using NLTK \citep{nltk}.
The bigram similarity measures and the $\lstm$ measure
improve BLEU scores for almost all beam sizes.
In contrast, diverse decoding has almost no effect on BLEU,
while length normalisation performs worse than standard beam search.
The best result with our similarity metrics is achieved by $\lstm$ at $k{=}100$.
This is significantly better than the best result for standard beam search ($k{=}10$),
with $p{<}0.001$ for a paired bootstrap test following \citet{koehn2004statistical}.
Using $\smoothbleu$ and increasing the beam size to~$k{=}100$ gives the overall best results.

Sampling methods proposed for open-ended generation perform poorly. 
Top-k sampling \citep{fan-etal-2018-hierarchical} achieves BLEU scores of 17.15 $(k{=}4)$ and 13.79 $(k{=}10)$,
nucleus sampling \citep{holtzman2020curious} achieves a score of 13.62 $(\textrm{top\_p}{=}0.9)$

Consistent with \citet{ott2018} and \citet{koehn2017six},
increasing~$k$ with beam search too much reduces BLEU.
However, this drop does not occur for our voting method.

\begin{table*}
	\centering	
	\begin{tabular}{l r r r | r r r | r r r} 
		& \multicolumn{3}{c}{Distinct captions}  & \multicolumn{3}{c}{Distinct unigrams} & \multicolumn{3}{c}{Distinct bigrams} \\
		Beam size $k$ & 
		\multicolumn{1}{c}{2} & \multicolumn{1}{c}{10} & \multicolumn{1}{c}{100} & 
		\multicolumn{1}{c}{2} & \multicolumn{1}{c}{10} & \multicolumn{1}{c}{100} & 
		\multicolumn{1}{c}{2} & \multicolumn{1}{c}{10} & \multicolumn{1}{c}{100} \\
		\hhline{==========}
		Beam search  & 
		9208 & 5488 & 4150  & 
		668 & 621 & 605 &
		3395 & 2778 & 2479 \\
		\hdashline
		Length normalisation & 
		9978 & 6418 & 5039 & 
		681 & 627 & 587 &
		3502 & 2863 & 2471 \\
		Diverse decoding & 
		9942 & 6424 & 4403 & 
		672 & 646 & 612 &
		3402 & 3023 & 2561 \\
		\hdashline
		$\overl_1$  &
		\textbf{10727} & \textbf{8916} & \textbf{10808} & 
		\textbf{687} & \textbf{646} & 628 & 
		\textbf{3576} & 3232 & 3596 \\
		$\preci_1$  & 
		\textbf{10727} & 8902 & 10768 & 
		\textbf{687} & 645 & 638 & 
		3572 & \textbf{3238} & \textbf{3607} \\
		$\overl_2$  & 
		9519 & 7598 & 9221 & 
		673 & 620 & 580 &
		3446 & 2854 & 2887 \\
		$\preci_2$  & 
		9522 & 7590 & 9248 & 
		673 & 620 & 581 & 
		3444 & 2848 & 2892 \\
		$\lstm$  & 
		9208 & 7613 & 10133 &
		668 & 629 & \textbf{655} &
		3395 & 2891 & 3331 \\
		\hdashline
		$\bleu$ \small{(MBR)} &
		9159 & 6512 & 6763 &
		667 & 612 & 570 &
		3392 & 2666 & 2446 \\
		$\smoothbleu$ \small{(MBR)} &
		9206 & 6522 & 7019 &
		667 & 613 & 560 &
		3396 & 2675 & 2415 \\
	\end{tabular}%
	\caption{%
		Number of distinct captions, unigrams and bigrams in the generated captions.
	}
	\label{table/caption_diversity}
\end{table*}

\subsubsection{Caption length}
\label{sec:length}

\begin{table}
	\centering	
	\begin{tabular}{l r r r r}
		& \multicolumn{4}{c}{Average caption length}\\
		Beam size $k$ & \multicolumn{1}{c}{1} & \multicolumn{1}{c}{2} & \multicolumn{1}{c}{10} & \multicolumn{1}{c}{100} \\
		\hline
		\hline
		Beam search & 8.41 & 8.79 & 9.18 & 9.11\\
		\hdashline
		Length norm. & 8.41 & 9.19 & 10.24 & 10.43\\
		Diverse decod. & 8.41 & 8.71 & 9.12 & 9.15\\
		\hdashline
		$\overl_1$ & 8.41 & \textbf{9.22} & \textbf{10.40} & \textbf{11.20}\\
		$\preci_1$ & 8.41 & 9.21 & 10.38 & 11.15 \\
		$\overl_2$ & 8.41 & 8.96 & 9.86 & 10.55 \\
		$\preci_2$ & 8.41 & 8.96 & 9.86 & 10.55 \\
		$\lstm$ & 8.41 & 8.79 & 9.17 &  8.82 \\
		\hdashline
		$\bleu$ \small{(MBR)} & 8.41 & 8.77 & 9.27 &  9.32 \\
		$\smoothbleu$ & 8.41 & 8.79 & 9.24 &  9.13 \\
		
	\end{tabular}%
	\caption{
		Average length of the generated captions.
		The reference captions contain on average 10.59 words.}
	\label{table/caption_length}
\end{table}

To analyse differences between methods,
we first look at caption length, shown in Table~\ref{table/caption_length}.
Standard beam search produces slightly longer captions as $k$~increases up to~10.
All n-gram measures generate longer captions than standard beam search,
and length continues to increase as $k$~goes to~100.
Length normalisation also increases caption length,
but this is at the cost of BLEU score (see~\S\ref{sec:bleu}).
Diverse decoding does not increase caption length.
The $\lstm$ measure produces slightly shorter captions
-- as it is symmetric, it does not favour long sequences as the asymmetric n-gram measures do (see~\S\ref{sec:vote}).
As predicted by our range voting interpretation, 
MBR, for which the asymmetry is in the opposite direction,
produces shorter captions than the simple n-gram similarity metrics.

\subsubsection{Caption diversity}
\label{sec:diversity}

Following the approach of \citet{li2015diversity},
\citet{dhingra2016towards}, and
\citet{xu2017neural,xu2018better},
we investigate the diversity of the generated captions
by counting the number of distinct captions, unigrams, and bigrams (see Table~\ref{table/caption_diversity}).

For standard beam search, the number of distinct captions drops as $k$~increases.
Both baselines weaken this effect, but the drop is still present.
In contrast, range voting maintains caption diversity as $k$~increases,
for all similarity measures.

Similarly, standard beam search sees a drop in the number of distinct unigrams and bigrams as $k$~increases,
and the baselines do not seem to mitigate this.
In contrast, the unigram measures and the $\lstm$ measure
maintain both unigram diversity and bigram diversity as $k$~increases,
while the bigram measures partially maintain bigram diversity.
As expected from our range voting perspective, MBR generates less diverse captions.

\subsubsection{Human evaluation}
\label{sec:human}
BLEU is known to be imperfect,
and does not always match human judgements \citep{callison2006bleu,blain2017exploring}.
While the n-gram similarity measures produce similar BLEU scores to standard beam search,
they also produce longer captions,
which are potentially more informative.
To investigate whether they are more informative in way that is not reflected by BLEU,
we took 500 validation images for human evaluation,
comparing the captions produced by standard beam search ($k{=}10$)
against our best-performing n-gram measure ($\preci_2$, $k{=}100$).
Each pair of captions was presented in a random order, with the original image,
and judged on a five-point scale (one caption much better, slightly better, or no difference).

The voted caption was rated better 106 times, and worse 73 times.
This is statistically significant, with $p{=}0.0165$ for a two-tailed sign test,
discarding ties \citep{emerson1979sign}.
However, for captions rated much better,
the voted caption was better 27 times and worse 40 times.
This is suggestive but not fully significant ($p{=}0.142$).

These results support the claim that a voted caption
represents more of the information present in a model's distribution over captions
-- this often leads to a better caption,
but where the model is wrong, adding wrong information can make the caption much worse.
After all, our method is designed as a better way to select from a distribution,
not as an improvement to the distribution itself.

\begin{table*}[!ht]
\centering	
\begin{tabular}{l r r r r r r} 
Beam size $k$ & 
	\multicolumn{1}{c}{1} & 
	\multicolumn{1}{c}{2} & 
	\multicolumn{1}{c}{4} &
	\multicolumn{1}{c}{10} &
	\multicolumn{1}{c}{30} &
	\multicolumn{1}{c}{100} \\
\hhline{=======}
Beam search 
	& 24.04 & 25.10 & 25.36 & 24.91 & 23.46 & 20.56  \\
\hdashline
Length normalisation 
	& 24.04 &  25.19 & \textbf{25.59} &  25.55 &  24.40 & 21.78 \\
Diverse decoding
    & 24.04 & 24.88 & 25.17 & 24.71 & 23.49 & 20.82 \\
Diverse beam search
	& 24.04 & 24.55 & 24.70 & 23.93 & 22.14 & 18.38 \\
Beam search (no copy)
	& 23.96 & 25.10 & 25.43 & 25.23 & 24.38 &  22.59 \\
\hdashline
$\overl_1$ 
	& 23.96 & 25.17 & 25.48 & 25.55 & 24.97 & 24.20 \\
$\preci_1$ 
	& 23.96 & 25.17 & 25.47 & 25.54 & 24.95 & 24.21 \\
$\overl_2$ 
	& 23.96 & 25.14 & 25.49 & \textbf{25.70} & \textbf{25.08} & \textbf{24.62} \\
$\preci_2$ 
	& 23.96 & \textbf{25.20} & 25.53 & 25.39 & 24.69 & 23.96 \\
$\transformer$
	& 23.96 & 25.10 & 25.44 & 25.51 & 24.67 & 23.36 \\
\hdashline
$\bleu$ \small{(MBR)}
	& 23.96 & 25.09 & 25.42 & 25.51 & 24.79 & 23.53 \\
$\smoothbleu$ \small{(MBR)}
	& 23.96 & 25.10 & 25.42 & 25.51 & 24.81 & 23.65 \\
\end{tabular}%

\caption{%
	BLEU scores on newstest2014,
	with range voting applied to the beams obtained with no-copy filtering.
}
\label{table/ende_bleu}
\end{table*}

\subsection{Machine translation}
\label{sec:mt}

For the translation task, we use the WMT'14 English-German dataset,
consisting of 4.5M sentence pairs.
We train a Transformer `big' model \citep{vaswani2017attention},
implemented in the Tensor2Tensor library \citep{tensor2tensor}.
We use the joint source and target byte-pair encoding 
vocabulary \citep{sennrich2016neural} with 32,000 tokens
available on Tensor2Tensor. 
All results reported are for the newstest2014 test set, 
containing 2737 sentence pairs \citep{bojar2014wmt}.\footnote{%
	We are evaluating systems translating from English into German,
	but half of the newstest2014 sentences were originally in German and translated into English.
	Translation artifacts are known to have an impact on
	machine translation performance
	(for example: \citealp{kurokawa2009automatic,holmqvist2009augment,lembersky2012translationese}).
	One reviewer asked whether there is a difference in performance
	for the two halves of the dataset,
	as found by \citet{freitag2019ape}.
	In terms of BLEU score, range voting appears more effective
	for forward-translation (original text in English),
	but in terms of manual evaluation, it appears more effective
	for backward-translation (original text in German).
	For reasons of space,
	we only report results for the whole dataset.
}
The BLEU scores were computed using
SacreBleu \citep{sacrebleu}.

\citet{ott2018} found that a common source of model error 
comes from outputting a copy of the input sentence,
still in the source language.
We also observe this phenomenon:
with beam size~4, 0.4\%~of the outputs are exact copies of the input.
This increases to 3.8\% of the outputs for beam size~100.
When counting the number of partial copies\footnote{%
	A partial copy is defined to be a generated sentence 
	containing at least 50\% of the unigrams in the input sentence.
} the effect is even stronger:
for beam sizes 4 and~100, respectively 1.3\% and 12.4\% 
of the generated translations are partial copies. 
Because of this, we add the method proposed by \citet{ott2018},
which filters out partial copies during beam search,
as an extra baseline.

\begin{figure*}
	\begin{center}
		\includegraphics[scale=0.24]{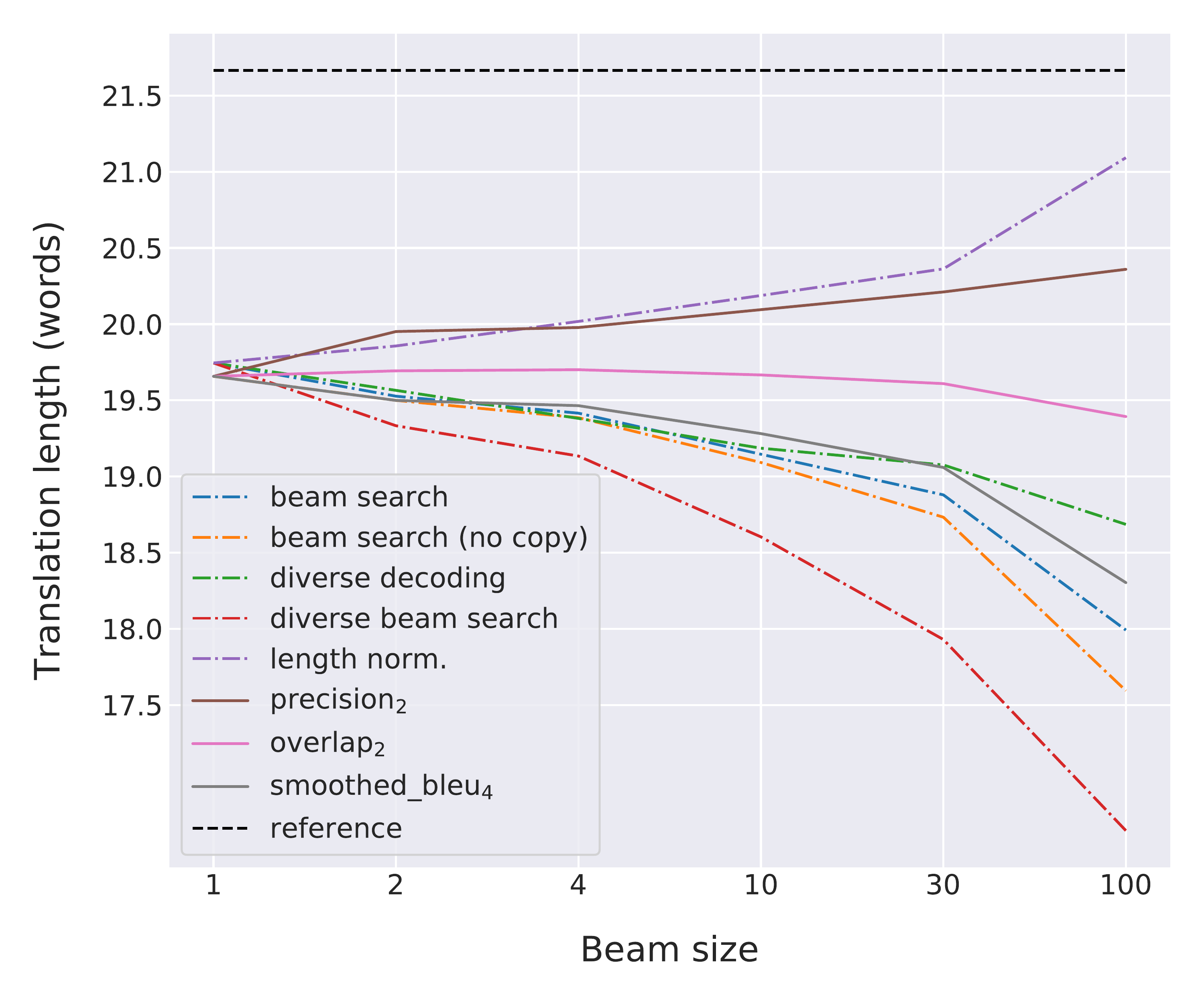}
		\hspace{8mm}
		\includegraphics[scale=0.24]{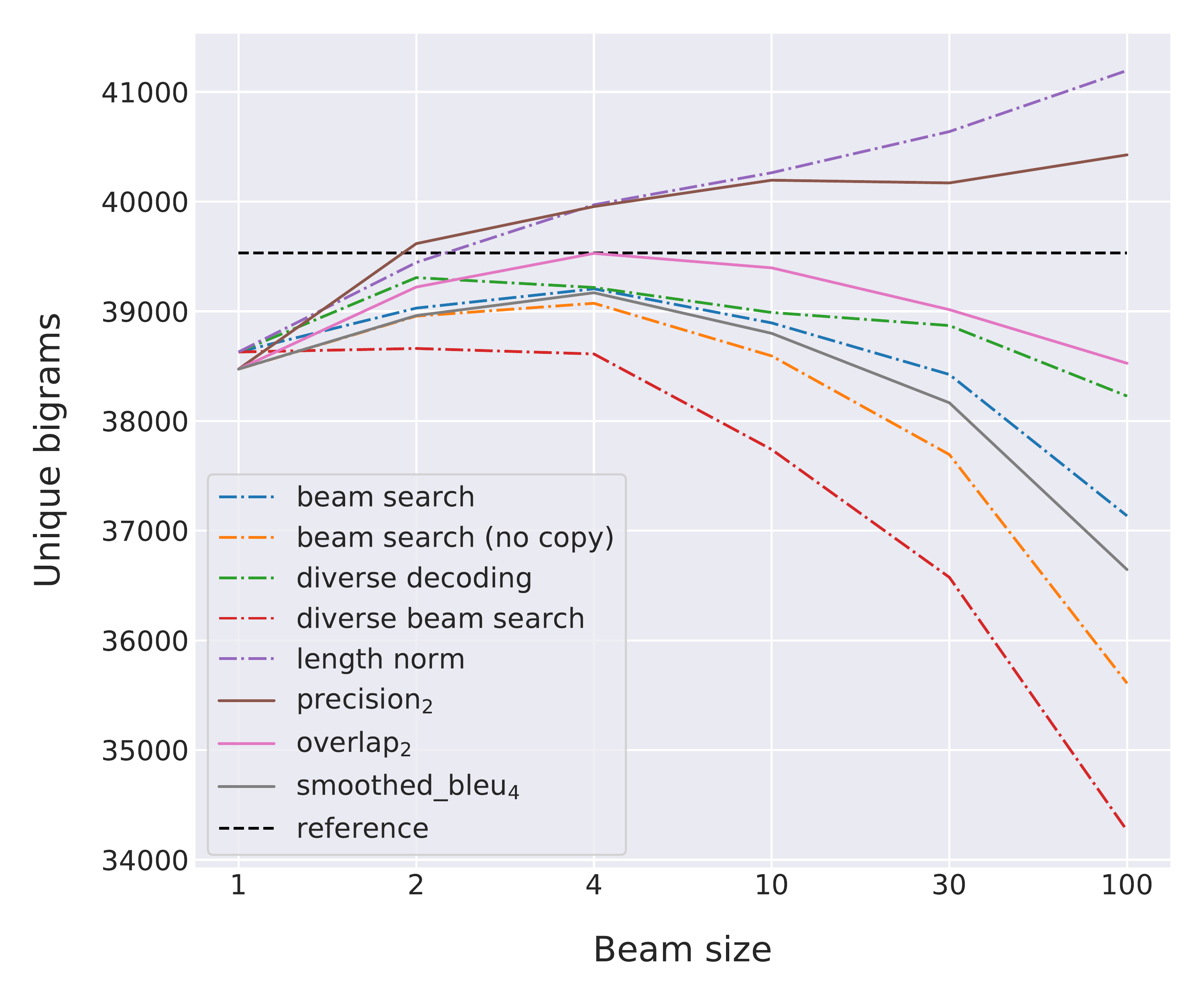}
	\end{center}
	\caption{%
		The lengths (left) and the number of unique bigrams in the generated translations (right).
		Baseline methods are shown as dashed lines, voting and MBR results as solid lines, and reference translations in black (the horizontal line).
		Full tables of results are given
		in Appendix~\ref{app/full_translation_stats}.
	}
	\label{fig:translation_stats}
\end{figure*}

\subsubsection{BLEU scores}
\label{sec:mt:bleu}

The BLEU scores obtained on the WMT'14 En-De newstest2014 test set 
are shown in Table~\ref{table/ende_bleu}.

For beam search and all considered baselines, the scores 
for the larger beam sizes drop considerably.
Adding the copy pruning heuristic from \citet{ott2018} 
does help mitigate this problem somewhat but does not solve it:
there is almost a 3 BLEU point drop between $k{=}4$ and~$k{=}100$.

To decouple a trivial source of model errors (input copies)
from search errors,
we apply our range voting method
on the beams obtained with the filtering heuristic
(Table~\ref{table/ende_bleu}, bottom half).
Regardless of which similarity metric is used, 
re-ranking using range voting improves the BLEU score,
and with the $\overl_2$ similarity, 
we achieve the best overall score of 25.70.
Furthermore, the performance drop at large beam sizes
is reduced when using range voting to about 1 BLEU point
for $\overl_2$.

There are two possible reasons for lower performance at larger beams: 
(1) different candidates:
the sentence selected for a small beam
is not in the larger beam;
or (2) different voter preferences:
the sentence selected for a small beam size is still there, but
range voting selects a different sentence.
In fact, both phenomena occur.
First, for beam search and all similarity metrics,
about 10\% and 5\% of the sentences
selected at~$k{=}4$ and~$k{=}10$ respectively 
are not in the beam of size 100.
Second, 48\% and 61\% of the sentences chosen by standard beam search 
with~$k{=}4$ and~$k{=}10$ respectively are also chosen for $k{=}100$,
but this drops to 32\% and 36\% respectively when using
range voting with $\overl_2$ similarity.
This suggests that generating candidates and voters
independently could lead to further improvements,
which we explore in~\S\ref{sec/more_voters}.

Sampling methods also perform poorly on this task. 
Top k sampling achieves BLEU scores of 17.39 $(k{=}4)$ and 15.21 $(k{=}10)$, 
nucleus sampling achieves a score of 10.10 $(\textrm{top\_p}{=}0.9)$

\subsubsection{Translation length}

The average length of the generated translations
are shown in Figure~\ref{fig:translation_stats}.
All similarity metrics generate longer translations than
standard beam search with and without filtering, 
but shorter than length normalisation.
At beam size~$k{=}100$, length normalised beam search 
generates almost an extra word per translation compared to~$k{=}30$.

Just as for image captioning, the length of translations
generated by standard beam search decreases as the beam size increases.
We again note that the translations generated by range voting 
with asymmetric similarity metrics are on average longer,
except for MBR where the asymmetry in the similarity 
metric penalises longer candidates.
However, it is no longer the case that increasing the beam size
also increases the length of the translations generated by range voting.

\subsubsection{Translation diversity}

The numbers of distinct bigrams generated
are shown in Figure~\ref{fig:translation_stats}.
Out of diverse decoding and diverse beam search, which aim to
increase diversity within a beam, only diverse decoding increases the number of generated bigrams
compared to beam search.
Length normalisation generates the most unique bigrams,
and this increases with beam size,
also due to the translations being longer on average.
On the other hand,
the copy filtering heuristic decreases 
the number of distinct bigrams generated.
Just as for image captioning, range voting increases the diversity
of the generated translations.
For all similarity metrics,
more unique bigrams are generated than 
beam search with copy filtering 
(on top of which range voting was applied).
Furthermore, the simple n-gram metrics generate 
more unique bigrams than standard beam search, 
recovering the drop occurring for the filtering heuristic.

\subsubsection{Human evaluation}
\label{sec:human-mt}

We used a human evaluation
to investigate differences not reflected by BLEU.
For 500 sentences, we compared
the strongest baseline (length normalisation, ~$k{=}4$)
with range voting ($\preci_2$,~$k{=}10$,
as this performed well on BLEU, length, and diversity),
following the procedure as in~\S\ref{sec:human}.
The voted translation was rated better 69 times, and worse 44 times.
This is statistically significant, with $p{=}0.0235$ for a two-tailed sign test.
For translations rated much better,
the difference is not significant (36 better, 28 worse).

\subsubsection{Including more voters}

The range voting formulation doesn't require the set of 
candidates~$\mathcal{C}$ and voters~$\mathcal{V}$ to be the same
(see Equation~\ref{eqn:score}). 
We can capture more knowledge from the underlying distribution 
by using a larger and more diverse set of voters 
(and could be acquired more efficiently by repeatedly sampling)
whilst constraining the set of candidates to avoid model errors.
This was similarly done by \citet{tromble2008lattice},
who refer to the sets of voters and candidates
as the ``evidence'' and ``hypothesis'' spaces.

For the voters, we increase~$k$ from 4 to 1000
and apply 3 different search methods:
sampling~$k$ times, 
stochastic beam search \citep{kool2019stochastic},
and beam search with copy filtering.
For the candidates we use beam search with copy filtering and~$k{=}4$.
We fix the similarity metric to $\overl_2$,
which was the best performing metric for large~$k{\geq}4$ (~\S\ref{sec:mt:bleu}).

\begin{figure}
	\begin{center}
		\includegraphics[scale=0.24]{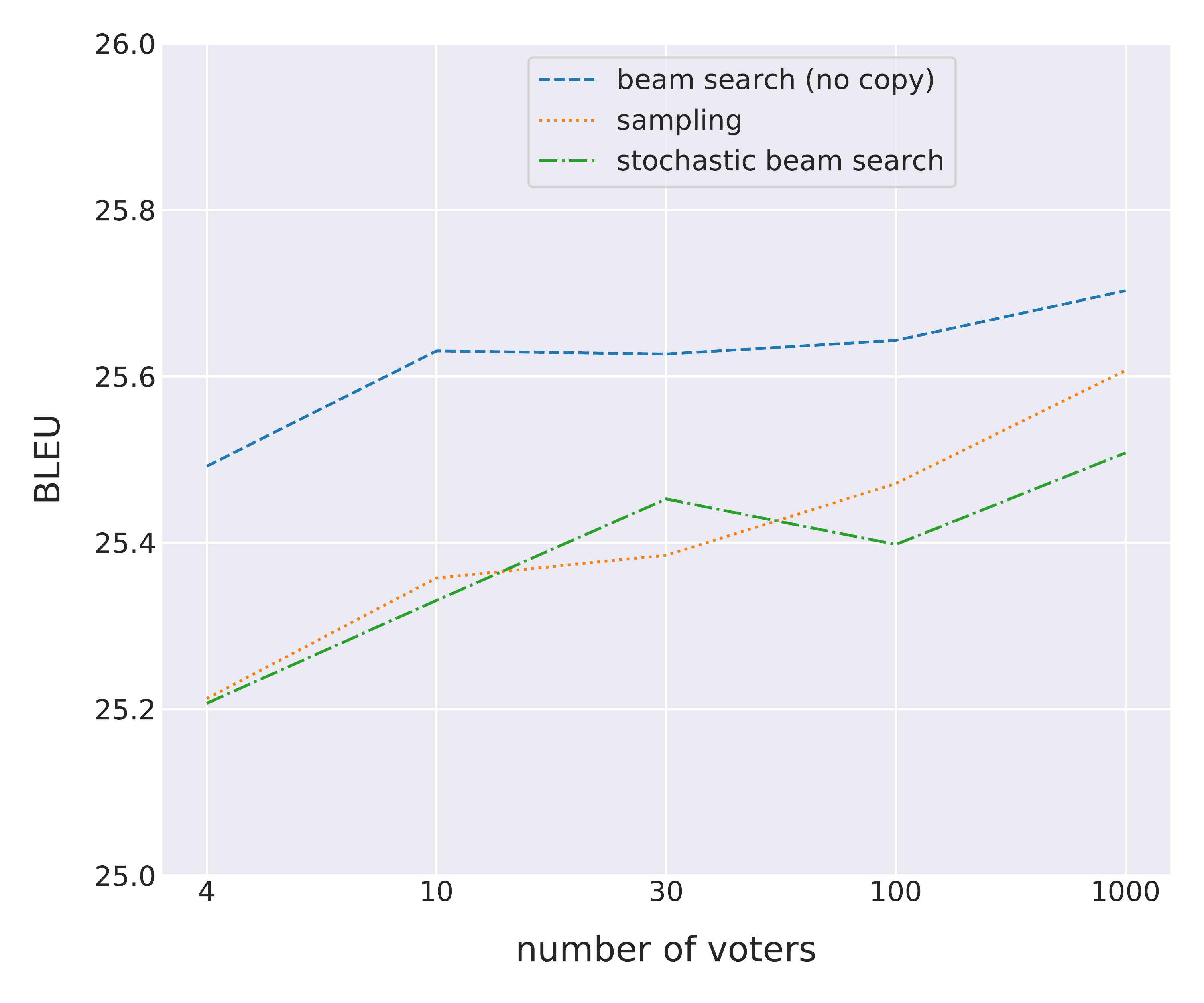}
	\end{center}
	
	\caption{%
		Performance of $\overl_2$ range voting,
		varying the number of voters for a fixed set of candidates.
	}
	\label{fig/bleu_asymmetric_voting}
\end{figure}

For all 3 generation methods, increasing the number of voters
increases BLEU (Figure~\ref{fig/bleu_asymmetric_voting}), 
suggesting that the previous drop in performance
is due to worse candidates in larger beams,
rather than worse voter preferences.

\label{sec/more_voters}

\section{Conclusion}
\label{sec:conc}
Instead of generating the most \textit{likely} sequence,
we propose a method to generate the most \textit{representative} sequence,
formalising representativeness using a similarity measure and range voting.

The evaluation on image captioning 
and machine translation shows that
despite using simple similarity measures,
we achieve an increase in BLEU score,
an increase in caption length and diversity,
and statistically significantly better human evaluation performance 
on both tasks.

For the image captioning task,
performance of our method does not drop 
as beam size increases,
removing the sensitivity of results to this hyperparameter.
On the machine translation task, 
performance does drop for larger beam sizes, 
although by much less than with standard beam search 
or the baselines.
Furthermore, performance increases
as the number of voters increases,
for a fixed set of candidates.

Using better similarity measures that capture semantics
could further improve results and is a promising direction
for further research.

Finally, our approach can be applied to 
any probabilistic language model,
without any need for additional training.
This opens up many other tasks,
including summarisation, dialogue systems, and question answering.
If multiple outputs can be used
(e.g.\ offering options to a user),
our method can be extended to use reweighted range voting \citep{smith2005rerange},
a procedure that elects multiple candidates.

\section*{Acknowledgements}

We would like to thank Kris Cao
for discussions about distributions over sequences,
which prompted the initial idea for this project.
We would like to thank Dr.\ Robert Harle and Prof.\ Ann Copestake
for making this project possible, and for providing some early feedback.
We would like to thank Andreas Vlachos,
Guy Aglionby, James Thorne, Chris Davis,
and the NLIP reading group in Cambridge,
for feedback on earlier drafts of this paper.
Finally, we would like to thank Chris Dyer
for his insightful comments and suggestions.

\bibliography{acl2020}
\bibliographystyle{acl_natbib}

\newpage
\appendix

\onecolumn

\section{Translation length and diversity}
\label{app/full_translation_stats}

\centering
\vfill

\begin{tabular}{l r r r r r r}
& \multicolumn{6}{c}{Average translation length}\\
Beam size $k$ 
	& \multicolumn{1}{c}{1} 
	& \multicolumn{1}{c}{2} 
	& \multicolumn{1}{c}{4} 
	& \multicolumn{1}{c}{10} 
	& \multicolumn{1}{c}{30} 
	& \multicolumn{1}{c}{100} \\
\hline
\hline
Beam search 
	& 19.74 & 19.53 & 19.42 & 19.15 & 18.88 & 17.99 \\
\hdashline
Length norm.
	& 19.74 & 19.86 & 20.02 & 20.19 & 20.36 & 21.09 \\
Diverse decoding 
	& 19.74 & 19.57 & 19.38 & 19.19 & 19.08 & 18.69 \\
Diverse beam search
	& 19.74 & 19.33 & 19.13 & 18.60 & 17.93 & 16.68 \\
Beam search (no copy)
	& 19.66 & 19.50 & 19.39 & 19.09 & 18.73 & 17.59 \\
\hdashline
$\overl_1$ 
	& 19.66 & 19.79 & 19.86 & 19.93 & 20.03 & 19.96 \\
$\preci_1$ 
	& 19.66 & 19.79 & 19.86 & 19.93 & 20.03 & 19.96 \\
$\overl_2$ 
	& 19.66 & 19.69 & 19.70 & 19.67 & 19.61 & 19.39 \\
$\preci_2$ 
	& 19.66 & 19.95 & 19.98 & 20.09 & 20.21 & 20.36 \\
$\transformer$
    & 19.66 & 19.50 & 19.48 & 19.31 & 19.12 & 18.27 \\
\hdashline
$\bleu$ \small{(MBR)}
	& 19.66 & 19.50 & 19.46 & 19.28 & 19.06 & 18.18 \\
$\smoothbleu$ \small{(MBR)}
	& 19.66 & 19.50 & 19.46 & 19.28 & 19.06 & 18.30 \\
\end{tabular}%
\captionof{table}{
	Average length of the generated translations on the newstest2014 dataset. 
	The reference translations contain on average 21.66 words, more than for any of the above generation methods.}

\vfill

\begin{tabular}{l r r r r r r } 
& \multicolumn{6}{c}{Number of distinct bigrams}\\
Beam size $k$ 
	& \multicolumn{1}{c}{1} 
	& \multicolumn{1}{c}{2} 
	& \multicolumn{1}{c}{4} 
	& \multicolumn{1}{c}{10} 
	& \multicolumn{1}{c}{30} 
	& \multicolumn{1}{c}{100}  \\
\hhline{=======}
Beam search
	& 38629 & 39029 & 39205 & 38894 & 38424 & 37135 \\
\hdashline
Length normalisation 
	& 38629 & 39445 & 39971 & 40263 & 40638 & 41195 \\
Diverse decoding
	& 38629 & 39307 & 39217 & 38989 & 38870 & 38227 \\
Diverse beam search
	& 38629 & 38661 & 38611 & 37740 & 36575 & 34266 \\
Beam search (no copy)
	& 38473 & 38956 & 39073 & 38593 & 37694 & 35609 \\
\hdashline
$\overl_1$ 
	& 38473 & 39353 & 39743 & 39775 & 39636 & 39334 \\
$\preci_1$
	& 38473 & 39354 & 39739 & 39772 & 39634 & 39334 \\
$\overl_2$
	& 38473 & 39221 & 39528 & 39396 & 39015 & 38526 \\
$\preci_2$ 
	& 38473 & 39617 & 39955 & 40195 & 40170 & 40426 \\
$\transformer$
    & 38473 & 38956 & 39223 & 38930 & 38277 & 36787 \\
\hdashline
$\bleu$ \small{(MBR)}
	& 38473 & 38957 & 39157 & 38792 & 38166 & 36564 \\
$\smoothbleu$ \small{(MBR)}
	& 38473 & 38961 & 39169 & 38799 & 38166 & 36645 \\

\end{tabular}%
\captionof{table}{%
	Number of distinct bigrams in the generated translations for the newstest2014 dataset.
	The reference translations consist of 39,533 unique bigrams.
}

\vfill

\end{document}